\documentclass[twocolumn]{IEEEtran}

\usepackage{paralist}
\usepackage{graphicx}
\usepackage{graphbox}
\usepackage{amsmath}
\usepackage{amssymb}
\usepackage{tabulary}
\usepackage{subfigure}
\usepackage{multirow}
\usepackage{makecell}
\usepackage[backref]{hyperref}
\usepackage[numbers,sort&compress]{natbib}
\usepackage[usenames,dvipsnames]{color}
\usepackage[a4,center]{crop}
\usepackage{caption}
\usepackage{type1cm}
\usepackage{mathdots}
\usepackage{amsfonts}
\usepackage{amssymb}
\usepackage{pifont}
\usepackage{bbding}
\usepackage{utfsym}

\newcommand{\blist}{\begin{list}\setlength{\topsep 0pt \parsep 0pt}}
\newcommand{\elist}{\end{list}}
\newtheorem{thm}{Theorem}
\newcommand{\bthm}{\begin{thm}\begin{textit}}
\newcommand{\ethm}{\end{textit}\end{thm}}
\newtheorem{dfn}{Definition}
\newcommand{\bdfn}{\begin{dfn}\begin{textit}}
\newcommand{\edfn}{\end{textit}\end{dfn}}
\newtheorem{obn}{Observation}
\newcommand{\bobn}{\begin{obn}\begin{textit}}
\newcommand{\eobn}{\end{textit}\end{obn}}

\newtheorem{lema}{Lemma}
\newcommand{\bdla}{\begin{lema}\begin{textit}}
\newcommand{\edla}{\end{textit}\end{lema}}
\newtheorem{pro}{Property}
\newcommand{\bpro}{\begin{pro}\begin{textit}}
\newcommand{\epro}{\end{textit}\end{pro}}

\newcounter{algnum1}

\newcounter{algnum2}

\newcounter{algnum3}

\newcommand\bit{\begin{itemize}}
\newcommand\eit{\end{itemize}}

\newcommand{\Fig}[1]{Fig.~\ref{#1}}

\newcommand{\Sec}[1]{Section~\ref{#1}}
\newcommand{\Tab}[1]{Table~\ref{#1}}

\newcommand{\balgo}{
   \begin{description}
   \parskip=0pt
   \topsep=0pt
}
\newcommand{\ealgo}{
   \end{description}
}

\graphicspath{{img/}}

\begin{document}

\title{Semi-Self Representation Learning for Crowdsourced WiFi Trajectories}
\author{ Yu-Lin Kuo\IEEEauthorrefmark{1},
         Yu-Chee Tseng\IEEEauthorrefmark{1},
         Ting-Hui Chiang\IEEEauthorrefmark{2},
         Yan-Ann Chen\IEEEauthorrefmark{3}\\
		\IEEEauthorrefmark{1} Department of Computer Science, National Yang Ming Chiao Tung University, Taiwan\\
        \IEEEauthorrefmark{2} Advanced Technology Laboratory, Chunghwa Telecom Laboratories, Taiwan\\
        \IEEEauthorrefmark{3} Department of Computer Science and Engineering, Yuan Ze University, Taiwan
}
\maketitle

\begin{abstract}\label{Sec:abs}
WiFi fingerprint-based localization has been studied intensively. Point-based solutions rely on position annotations of WiFi fingerprints. Trajectory-based solutions, however, require end-position annotations of WiFi trajectories, where a WiFi trajectory is a multivariate time series of signal features. A trajectory dataset is much larger than a pointwise dataset as the number of potential trajectories in a field may grow exponentially with respect to the size of the field.
This work presents a semi-self representation learning solution, where a large dataset $C$ of crowdsourced unlabeled WiFi trajectories can be automatically labeled by a much smaller dataset $\tilde C$ of labeled WiFi trajectories. The size of $\tilde C$ only needs to be proportional to the size of the physical field, while the unlabeled $C$ could be much larger.
This is made possible through a novel ``cut-and-flip'' augmentation scheme based on the meet-in-the-middle paradigm. 
A two-stage learning consisting of trajectory embedding followed by endpoint embedding is proposed for the unlabeled $C$.
Then the learned representations are labeled by $\tilde C$ and connected to a neural-based localization network.
The result, while delivering promising accuracy, significantly relieves the burden of human annotations for trajectory-based localization.
\end{abstract}


\section{Introduction}\label{Sec:intro}
Location-based service (LBS) is essential in smart city, smart driving, and tour guiding applications. The core of LBS is localization. Indoor localization techniques can be divided into three categories: geometrical positioning, pedestrian dead-reckoning (PDR), and fingerprint matching \cite{8586939}. Geometric approaches may be derived by triangulation, angle of arrival, and time difference of arrival of specific wireless signals. 
PDR does not require auxiliary signals, but only measures relative motions and thus may suffer from accumulative errors. Fingerprint-matching approaches require annotated fingerprint datasets \cite{electronics10010002, 10.1145/3372224.3380894}, but acquiring the datasets is costly. 
    
    
This work focuses on fingerprint-based solutions. While fingerprints collection by humans or robots is easy, assigning labels to fingerprints is laborious and error-prone. To address this issue, \cite{s18103419} uses robots for automatic data and label collection.
Inpainting and interpolation are addressed in \cite{8981805}. With the popularity of mobile devices, crowdsourcing is promising for data collection. However, such data are typically unlabeled and even unreliable. A framework for evaluating crowdsourced data quality is in \cite{8320769}. Estimating WiFi APs' propagation parameters with crowdsourcing is explored in \cite{https://doi.org/10.1049/el.2015.1724}, and a crowdsourcing-based SLAM mechanism for radio map construction is explored in \cite{Enabling}. Using crowdsourced data effectively in the deep learning domain remains a challenge.


WiFi-based localization can be either point-based (using a single fingerprint) or trajectory-based (using a sequence of fingerprints). This paper focuses on learning representations from WiFi trajectories, modeled as multivariate time series along roaming paths. We assume access to two datasets: a large unlabeled, crowdsourced set $C$, and a smaller labeled set $\tilde C$, with $|C| \gg |\tilde C|$ to reduce annotation effort. Our semi-self-supervised framework learns representations from $C$ in an unsupervised manner, then uses $\tilde C$ to assign pseudo-labels to $C$. We show that combining $C$ and $\tilde C$ improves localization accuracy, provided $C$ is sufficiently large and $\tilde C$ scales with the field size.

The kernel of our design is a ``cut-and-flip'' augmentation scheme following the meet-in-the-middle paradigm.
Specifically, given a long trajectory, if we cut it in the middle and flip the second half, then the resulting two sub-trajectories should meet in the middle position, meaning that their endpoint embeddings should be very close to each other.
If we repeat the same process many times, a lot of sub-trajectories shall meet in the same middle position,
and if any one of these sub-trajectories appears in $\tilde C$, all these trajectories will have an opportunity to converge to very similar embeddings. Therefore, those pseudo-labels would become meaningful, making localization possible.
Following this, we separate our learning into trajectory embedding and endpoint embedding. 
With the assistance of $\tilde C$, we can attain a higher localization accuracy. 
The results significantly reduce human annotation efforts.

\begin{figure}[!t]
    \centering
    \includegraphics[width=0.9\columnwidth]{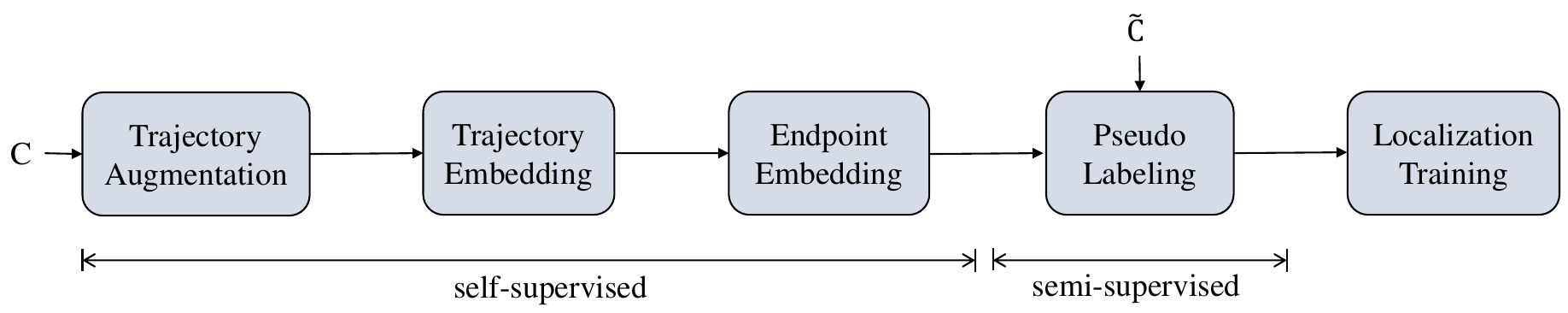}
    \caption{Workflow of our semi-self representation learning. Trajectories in crowdsourced $C$ are unlabeled, while those in $\tilde C$ are labeled, $|C| \gg |\tilde C|$.}
    \label{fig:Overall_1}
\end{figure}

In the literature, 
self-supervised learning (SSL) is proposed to extract intrinsic insights from data without labels. Predictive SSL makes a success in natural language processing \cite{devlin-etal-2019-bert,NEURIPS2020_1457c0d6,radford2019language,peters-etal-2018-deep,NIPS2017_3f5ee243}. 
Siamese networks use negative pairs \cite{10.5555/3524938.3525087, He_2020_CVPR}), while others use positive pairs only (e.g., BYOL \cite{10.5555/3495724.3497510}, SimSiam \cite{Chen_2021_CVPR} and SwAV \cite{NEURIPS2020_70feb62b}). 
Our framework will utilize positive pairs only because negative pairs are difficult to define and wireless signals in indoor environments may fluctuate significantly, resulting in numerous similar fingerprints for a location \cite{9758707}.
Our method is presented in \Sec{Sec:Method}. \Sec{Sec:exp_result} shows our experiment results. \Sec{Sec:conclusion} concludes this work.

\begin{figure}[!t]
    \centering
    \includegraphics[width=0.85\linewidth]{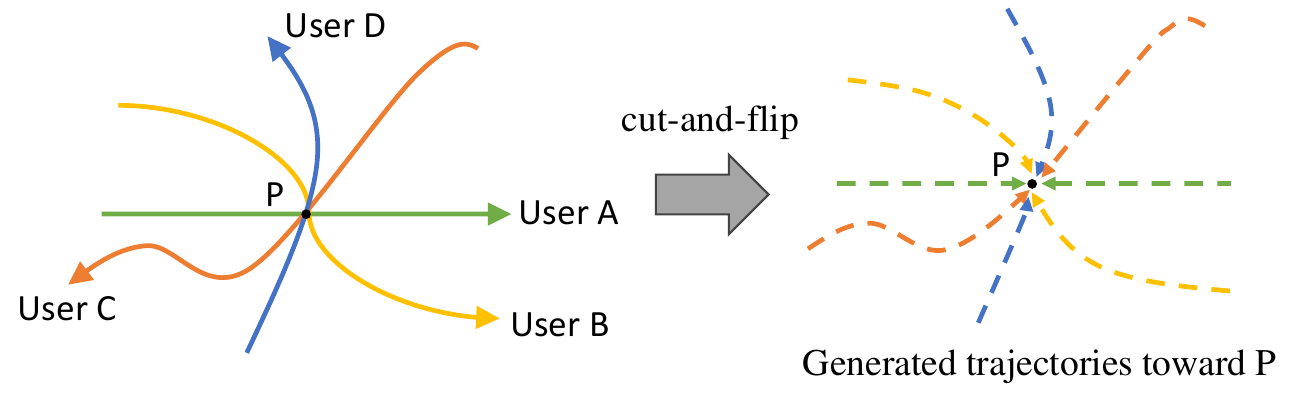}
    \caption{Cut-and-flip example.}
    \label{fig:reverse}
\end{figure}

\section{Semi-Self Representation Learning} 

\label{Sec:Method}

We are given a dataset $C$ containing a large number of crowdsourced WiFi trajectories that are unlabeled. A WiFi trajectory is a multivariate time series $X=[x_{i, j}]_{1:m, 1:t}$, where $m$ is the total number of observable WiFi APs, $t$ is the sequence length, and $x_{i, j}$ represents the features of the $i$th AP at the $j$th position in the trajectory, $1 \le i \le m$, $1 \le j \le t$.
On the other hand, a very small labeled dataset $\tilde C$ is given, $|\tilde C| \ll |C|$. Each WiFi trajectory in $\tilde C$ is accompanied by a position label in $R^2$, representing its end position. 

Our semi-self representation learning consists of two parts. The representation learning of the trajectories in $C$ is completely unsupervised. Then we assign pseudo labels to $C$ by utilizing the labels in $\tilde C$. The learning workflow is shown in \Fig{fig:Overall_1}. Trajectory augmentation first transforms the samples in $C$ into more samples, which are regarded as positive samples. Trajectory embedding and endpoint embedding then compute each trajectory's representation in a self-supervised manner. After pseudo labeling, the combined $C \cup \tilde C$ can be used for a localization task.

\subsection{Trajectory Augmentation}
\label{Sec:aug}
     

To deal with signal fluctuation, we design four ways to augment a WiFi trajectory for representation learning. 
Given a trajectory $X \in C$, we augment it to multiple $\tilde X$, regarded as positive samples of $X$. 

\begin{enumerate}
\item 
\textbf{Flipping:} This operation swaps $X$ on the temporal axis into $\tilde X$.
\item 
\textbf{Additive:} 
To model signal fluctuation, a small $\epsilon \sim \mathcal{N}(0, \alpha)$ is added to a randomly sampled $x_{i,j}$, resulting in the augmented sequence $\tilde{X}$, where $\alpha$ controls the noise magnitude.
\item \textbf{Scaling:}
Randomly sampled $x_{i,j}$ is scaled by a factor of $(1 + \epsilon)$, where $\epsilon \sim \mathcal{U}(-\beta, \beta)$ and $\beta$ is a tunable parameter, resulting in $\tilde{X}$.
\item \textbf{Masking:} 
To model intermittent WiFi signals, a short randomly sampled segment $[x_{i, j:j+s-1}]$ is regarded missing, where $1 \le j \le t$ and $s$ is a small integer. These missing signals are filled by the last or previously available signal $[x_{i, j-1}]$ or $[x_{i, j+s}]$. Note that multiple segments of $X$ can be masked to obtain $\tilde X$.
\end{enumerate}
	       
	        



In addition, we propose a new ``meet-in-the-middle'' paradigm to explore self representation learning. The augmentation is called \textbf{Cut-and-Flip}. Consider a length-$(2t-1)$ sequence $[X_{i, j}]_{1:m, 1:(2t-1)}$ in $C$.
We first cut it in the middle, resulting in two sequences $Z_1 = [X_{i, j}]_{1:m, 1:t}$ and $Z_2 = [X_{i, j}]_{1:m, t:(2t-1)}$.
We then flip $Z_2$ along the temporal axis, resulting in $\tilde Z_{2}$. Intuitively, if a person walks along $Z_1$ and another person walks along $\tilde Z_{2}$, they will meet in the same position (middle).
That is, $Z_1$ and $\tilde Z_{2}$ can be regarded as a positive pair, in terms of their end positions. 
In \Fig{fig:reverse}, we show an intersection point $p$ where many pedestrians pass through. All segments of the trajectory ending at $p$ and all segments of the trajectory departing from $p$, after flipped, can be regarded as positive samples. Therefore, when sufficient crowdsourced trajectories are collected, it is possible to self-learn similar representations from them without human annotations. This paradigm empowers crowdsourced WiFi trajectories to be used in localization tasks even if they are unlabeled.



\begin{figure}[!t]
    \centering
        \includegraphics[align=c,width=0.8 \linewidth]{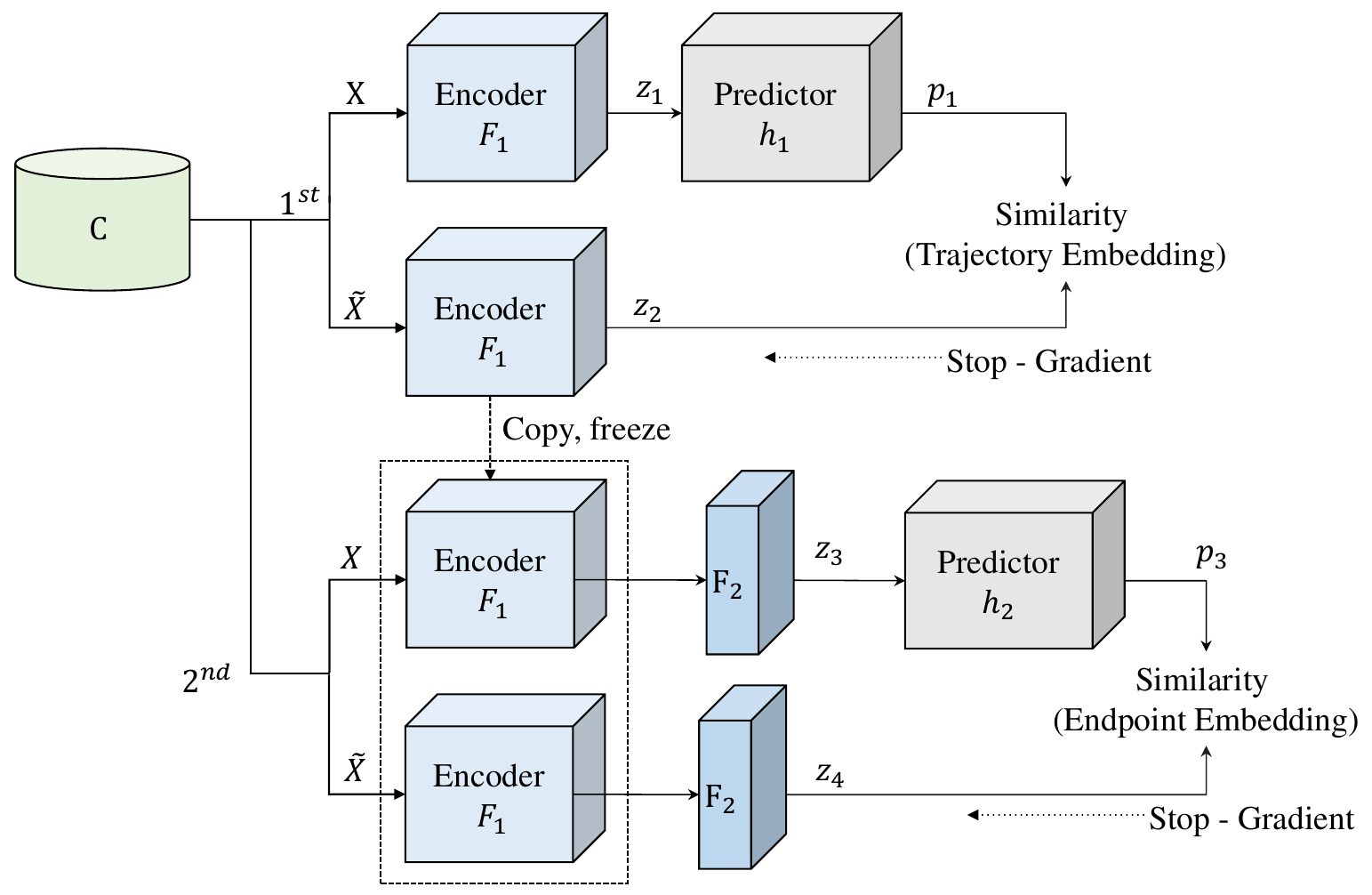}
    \caption{Two-stage representation learning.}
    \label{fig:SSL_phase}
\end{figure}

\subsection{Two-Stage Representation Learning}
\label{Sec:ssl-training}

With augmented positive pairs, we can perform representation learning. There are two stages: (i) The trajectory embedding will utilize pairs augmented by flipping, additive, scaling, and masking. 
(ii) The endpoint embedding will utilize pairs produced by cut-and-flip.
There are different frameworks for self-representation learning. 
As our WiFi trajectories' endpoints fall in a continuous physical space, negative pairs are hard to define. 
Hence, we choose to use SimSiam \cite{Chen_2021_CVPR}, which requires positive pairs only. 
The learning framework is shown in \Fig{fig:SSL_phase}, which has two pairs of Siamese networks.

\subsubsection{Trajectory Embedding}
The upper part of \Fig{fig:SSL_phase} is to learn the characteristics of WiFi trajectories. Given a trajectory $X \in C$ and its augmentation $\tilde X$, we consider them as a positive pair. Both $X$ and $\tilde X$ go through the encoder network $F_1$ and the predictor $h_1$.
    	\begin{equation}
    	    \label{eq:forward_1}
    	    z_1 = F_1(X),\; z_2 = F_1(\tilde X), \; p_1 = h_1(z_1), \; p_2 = h_1(z_2)
    	\end{equation}
We swap the roles of $X$ and $\tilde X$ and compute two negative cosine similarities: \begin{equation}
    \label{eq:negative_cosine_similarity}
    D(p_1, z_2) = -\frac{p_1}{||p_1||_2} \cdot \frac{z_2}{||z_2||_2}
\end{equation}
\begin{equation}
    \label{eq:negative_cosine_similarity}
    D(p_2, z_1) = -\frac{p_2}{||p_2||_2} \cdot \frac{z_1}{||z_1||_2}
\end{equation}
To avoid model collapse, stop-gradient (SG) is introduced. The loss function is defined as:
\begin{equation}
    \label{eq:loss}
    L = \frac{1}{2}D(p_1, SG(z_2)) + \frac{1}{2}D(p_2, SG(z_1)).
\end{equation}
The process is totally self-supervised. After finishing training, $F_1$ is frozen and used in the next stage.

\subsubsection{Endpoint Embedding}
The lower part of \Fig{fig:SSL_phase} is to learn the characteristics of the end positions of the WiFi trajectories. It has similar Siamese networks with a frozen $F_1$ copied from the first stage to maintain trajectory characteristics. Additional encoder $F_2$ and predictor $h_2$ are inserted. Given $X \in C$ and its augmentation $\tilde X$ (from cut-and-flip), we also consider them as a positive pair. Through $F_1$, $F_2$, and $h_2$, we obtain:
	    \begin{equation}
	    \begin{aligned}
	        \label{eq:f'}
	        &z_3 = F_2(F_{1}(X)), \;
            z_4 = F_2(F_{1}(\tilde X)), \\
            &p_3 = h_2(z_3), \;
            p_4 = h_2(z_4)
        \end{aligned}
	    \end{equation}
The similarity and loss functions are defined similarly. The embedding of the endpoint is written as $F_2(F_{1}(\cdot))$.

\subsection{Pseudo Labeling}
\label{Sec:pseudo-labeling}

\begin{figure}
    \centering
    \includegraphics[align=c,width=0.9 \linewidth]{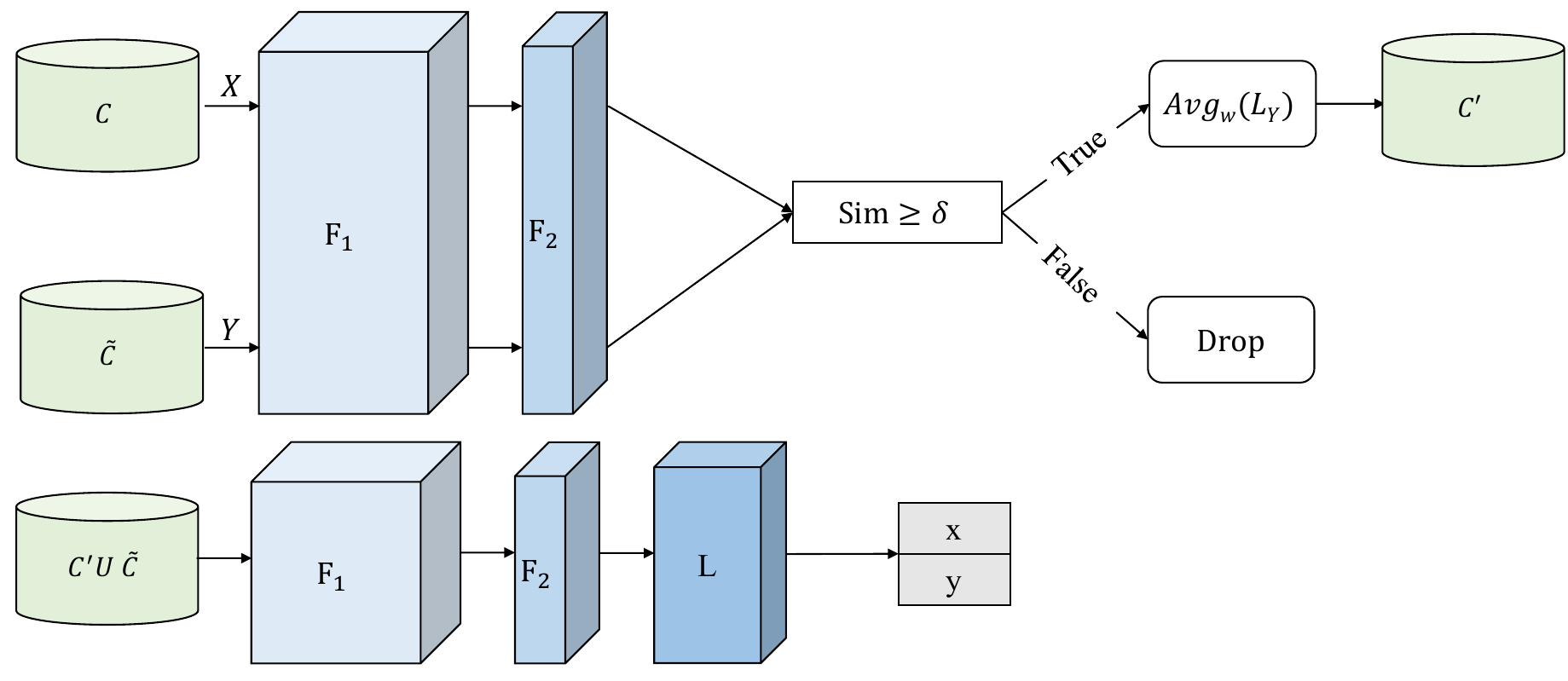}
    \caption{Top: pseudo labeling. Bottom: localization training.}
    \label{fig:Overall_2}
\end{figure}

With endpoint representations, we can conduct pseudo labeling on $C$ with the assistance of $\tilde C$. The workflow is shown in \Fig{fig:Overall_2}. For each $X \in C$, we compute its embedding $F_2(F_{1}(X))$ and compare it against $F_2(F_{1}(Y))$ for each $Y \in \tilde C$. If their cosine similarity is above a threshold $\delta$, the label of $Y$ is considered a candidate. If there are multiple candidates, a weighted average of these labels, $Avg_w(L_Y)$, is regarded as $X$'s label; otherwise, $X$ will be dropped. The refined dataset $C$ is named $C'$.

\subsection{Localization Training}
\label{Sec:model_training}

The size of the annotated set $\tilde C$ only needs to be proportional to the field size. 
However, the number of possible trajectories with respect to a filed may grow exponentially, making the crowdsourced set $C$ much larger than $\tilde C$. 
Therefore, the pseudo-labeled set $C'$ is potentially large too.

We use the labels of the union $\tilde C \cup C'$ to train a localization model, which is composed of $F_1$, $F_2$, and a mapping network $L$, as shown in \Fig{fig:Overall_2}. $F_1$ and $F_2$ help to find a WiFi trajectory's embedding that is unique to a position. Then $L$ maps the embedding to a position $(x, y)$. A lot of previous work has addressed the design of $L$, so we omit the details.




\section{Experiment Results}\label{Sec:exp_result}

\subsection{Datasets}
	    \label{Sec:dataset}
     
\begin{figure*}
    \centering
    \begin{minipage}{0.6\linewidth}
        \centering
        \includegraphics[width=0.47\linewidth]{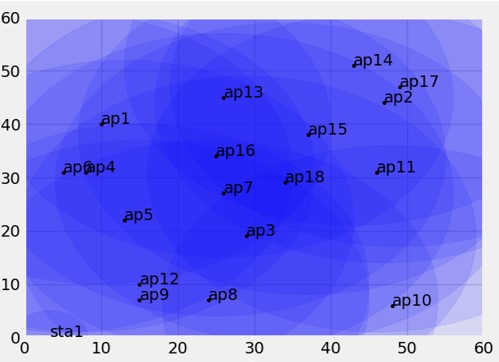}
        \hfill
        \includegraphics[width=0.47\linewidth]{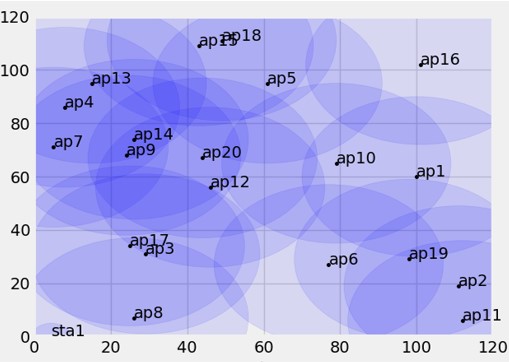}
        \caption{Simulation fields for Field 1 (left) and Field 2 (right).}
        \label{fig:Loc}
    \end{minipage}  
    \hfill
    \begin{minipage}{0.35\linewidth}
        \centering
        \captionof{table}{Training settings for $F_1$ and $F_2$.}
        \begin{tabular}{|c|c|}
            \hline
             learning rate &  0.01 \\ \hline
             initial decay epochs & 100 \\ \hline
             min decay learning rate & 0.0001 \\ \hline
             restart interval & 30 \\ \hline
             restart learning rate & 0.001 \\ \hline
             warmup epochs & 40 \\ \hline
             warmup start learning rate & 0.001 \\ \hline
        \end{tabular}
        \label{tab:coslr}
    \end{minipage}
\end{figure*}

To validate our claims, we construct two WiFi trajectory datasets using the emulation tool Mininet-WiFi \cite{10.1145/2934872.2959070, 7367387}. As illustrated in \Fig{fig:Loc}, we simulate two environments: Field 1 and Field 2, with dimensions of $60$m x $60$m and $120$m × $120$m, respectively. Each field contains 18 and 20 randomly deployed access points (APs), respectively.
The RSSI values are collected based on the log-normal shadowing propagation model. The path loss at a distance $d$ meters is determined by:
\begin{equation}
    \label{eq:propagationloss}
    P_L(d) = P_L(d_0) + 10n \log_{10}\left(\frac{d}{d_0}\right) + \chi
\end{equation}
where $P_L(d_0)$ is the path loss at a reference distance $d_0$, $n$ is the environment-dependent path loss exponent (set to $n = 4$), and $\chi \sim \mathcal{N}(0, \sigma^2)$ is a zero-mean Gaussian random variable representing shadowing effects, with standard deviation $\sigma = 1$. The maximum transmission range is set to $d_f = 30\,\mathrm{m}$. RSSI values are sampled on a 2D grid with $0.1$m spacing between adjacent points.



\begin{figure*}
    \centering
    \subfigure[Encoder $F_1$]{
        \label{fig:encoder}
        \raisebox{0.5\height}{
        \includegraphics[align=c,width=0.9\textwidth]{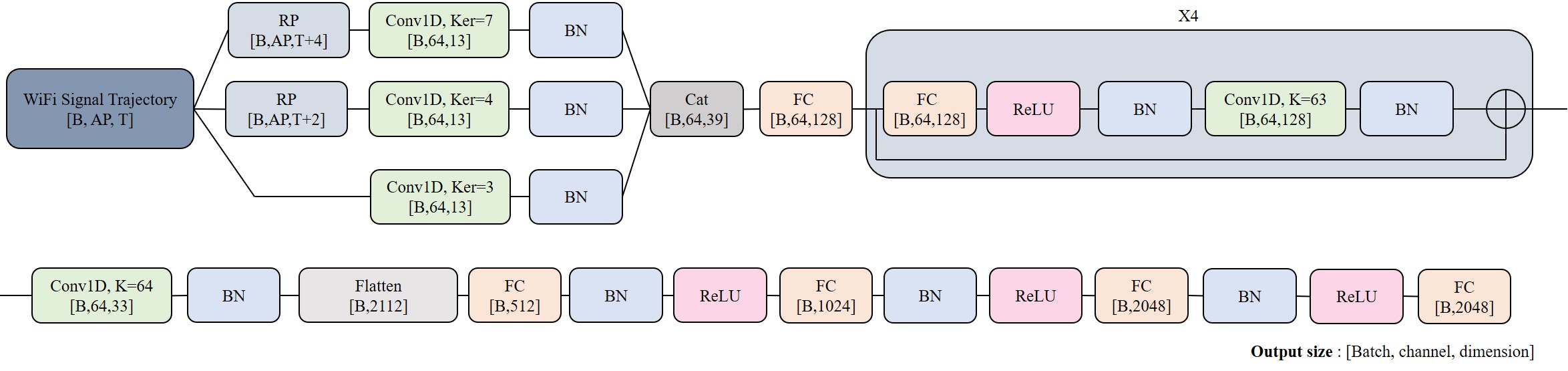}
        }
    }        	    
    \subfigure[Predictors $h_1$ and $h_2$]{
        \label{fig:predictor_h}
        \raisebox{0.5\height}{
        \includegraphics[align=c,width=0.28\textwidth]{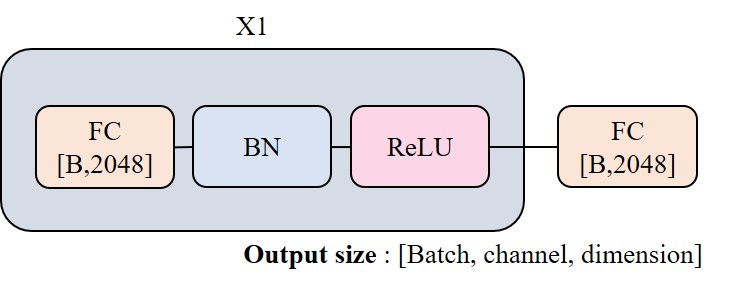}
        }
    }
    \subfigure[Encoder $F_2$]{
        \label{fig:projection_p}
        \raisebox{0.5\height}{
            \includegraphics[align=c,width=0.28\textwidth]{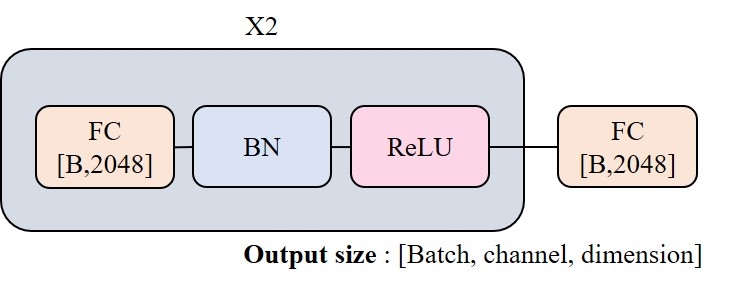}
        }
    }
    \subfigure[Model $L$]{
        \label{fig:Model-L}
        \raisebox{0.5\height}{ \includegraphics[align=c,width=0.28\textwidth]{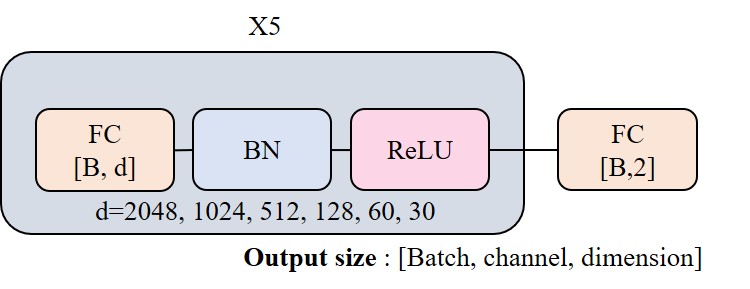}
        }
    }
    \caption{Model Structures. $B$=batch size, $AP$=Number of APs, $T$=sequence length, $RP$=Reflection Padding, $BN$=Batch Normalization, and $FC$=Fully Connected layer.
}
    \label{fig:model}
\end{figure*}    	    

To simulate human movement, we generate trajectories in $C$ and $\tilde{C}$ by connecting grid points using the Levywalk model \cite{5750071}, with speeds between $0.7$–$1.3$ m/s. Each trajectory lasts $T = 15$ seconds with 1-second intervals, yielding 15-point sequences. RSSI at each point is from its nearest grid location. We generate $1$M trajectories for the $C$ and $0.5$K labeled trajectories for $\tilde{C}$ in both fields. All RSSI values are distorted with noise $\epsilon \sim \mathcal{N}(0, 9)$ and normalized to $[0,1]$.

\subsection{Model Specifications}
\label{Sec:config}

Our model consists of a trajectory encoder $F_1$, an endpoint encoder $F_2$, two predictors ($h_1$, $h_2$), and a localization module $L$, as shown in \Fig{fig:model}. All components follow a projection-based architecture with 1D convolutions, batch normalization, ReLU activations, and fully connected (FC) layers. For first-stage trajectory augmentation, we apply the following configurations: (1) Additive: $\epsilon \sim \mathcal{N}(0, 0.2)$, (2) Scaling: $\epsilon \sim \mathcal{U}(-0.1, 0.1)$, (3) Masking: masked segment length $s \sim \mathcal{U}(3, 9)$ (20–60\% of each trajectory). We train $F_1$ using SGD with a cyclic cosine decay schedule (in \Tab{tab:coslr}) on an NVIDIA RTX 3090. After freezing $F_1$, we train $F_2$ using the same settings and then train $L$ with the Adam optimizer.

\subsection{Ablation Study}
        \label{sec:EXP1}  

\subsubsection{1-to-1 Embedding Distance} 
            \label{sec:EXP1-compare}

Given two crowdsourced trajectories $X_1$ and $X_2$ with the same endpoint, let $t_{X_1}$ and $t_{X_2}$ be their trajectory embeddings from the first-stage encoder $F_1$, and $e_{X_1}$ and $e_{X_2}$ be their endpoint embeddings from the second-stage encoder $F_2$. Let $s_t$ and $s_e$ denote the similarities between the respective pairs. Since both trajectories converge at the same endpoint, we expect $s_e \ge s_t$.

In Field 1, this condition holds in $67.1\%$ of cases. Among them, $19.7\%$ show a minor improvement ($< 0.1$), and $21.3\%$ show a significant improvement ($\ge 0.3$). In the remaining $32.9\%$ ($s_e < s_t$), most already have high similarity where $29.4\%$ fall within $[0.7, 1]$, and only $3.4\%$ fall below 0.7. Field 2 shows a similar trend, though the portion of $s_e \ge s_t$ is lower, likely due to reduced WiFi AP density. Nevertheless, only $8.6\%$ of non-improving cases fall below the $0.7$ threshold. These results validate the effectiveness of our two-stage framework. 

\begin{table}[!t]
        \centering
        \caption{Position error of pseudo labels by varying $\delta$.}
        \begin{tabular}{|c|c|c|c|c|}
            \hline
            $\|C\|$& \makecell{Similarity \\Threshold $\delta$} & \makecell{$\%$ of \\No labels} & CDF68 & CDF95 \\ \hline
            \multirow{2}{*}{\makecell{0.5k \\(Field 1)}}& 0.8 & 0.7\% & 3.19& 6.58\\ \cline{2-5} 
            & 0.9 & 9.8\% & 2.84& 5.39\\ \hline
            \multirow{2}{*}{\makecell{1k \\ (Field 1)}}& 0.8 & 0.01\% & 2.91& 6.36\\ \cline{2-5}
            & 0.9 & 3.3\% & 2.45& 4.93\\ \hline

            \multirow{2}{*}{\makecell{0.5k \\(Field 2)}}& 0.8 & 0.2\% & 10.0& 20.85\\ \cline{2-5} 
            & 0.9 & 3.8\% & 8.35& 18.12\\ \hline
            \multirow{2}{*}{\makecell{1k \\ (Field 2)}}& 0.8 & 0\% & 9.68& 20.35\\ \cline{2-5}
            & 0.9 & 1.2\% & 7.69& 17.36\\ \hline
        \end{tabular}
        \label{tab:pseudo-labeling}    
\end{table}

\begin{table*}
    \centering
    \caption{Comparisons of localization error on field 1. (Labeled dataset $\tilde C=0.5K$)}
    \scalebox{1.0}{
    \begin{tabular}{|c|c|c|c|c|c|c|c|c|c|c|}
        \hline
        Model & KNN & LSTM & NCP &  \multicolumn{7}{c|}{FULL}   \\ \hline
        
        Crowdsourced $C$& 0& 0& 0& 0& 10k& 200k& 10k& 200k& 10k& 200k\\ \hline
        
        Threshold $\delta$& - & - & - & - & \multicolumn{2}{c|}{0.8}& \multicolumn{2}{c|}{0.9} &\multicolumn{2}{c|}{0.99}\\ \hline
        
         Training dataset size& 0.5k& 0.6k & 0.5k & 0.5k& 10.5k& 199.8k& 9.6k& 181.2k& 1.4k& 9.8k \\ \hline
         
         Position error CDF68 & 7.81& 7.16& 8.37 & 6.7 & 5.77 & 4.7& 5.34& \textbf{4.16}& 5.97& 11.21 \\ \hline
         Position error CDF95 & 16.27& 13.14& 16.97 & 13.1 & 10.01 & 8.17& 8.61& \textbf{7.28}& 9.63& 20.06\\ \hline
    \end{tabular}
    }
    \label{tab:finetune}
\end{table*}             

\begin{table*}
    \centering
    \caption{Comparisons of localization error on field 2. (Labeled dataset $\tilde C=0.5K$)}
    \scalebox{1.0}{
    \begin{tabular}{|c|c|c|c|c|c|c|c|c|c|c|}
        \hline
        Model & KNN & LSTM& NCP& \multicolumn{7}{c|}{FULL}   \\ \hline
        
        Crowdsourced $C$& 0& 0& 0& 0& 10k& 200k& 10k& 200k& 10k& 200k\\ \hline
        
        Threshold $\delta$& -& -& -& -& \multicolumn{2}{c|}{0.8}& \multicolumn{2}{c|}{0.9} &\multicolumn{2}{c|}{0.99}\\ \hline
        
        Training dataset size& 0.5k& 0.5k & 0.5k & 0.5k& 10.5k& 199.8k& 9.6k& 193.2k& 1.4k& 57.5k \\ \hline
         Position error CDF68 & 22.95& 18.47& 59.19 & 57.81 & 14.98 & 11.53& 13.34& \textbf{10.32}& 18.64& 20.3 \\ \hline
         Position error CDF95 & 61.32& 35.84& 77.63 & 75.83 & 23.44 & 21.69& 22.21& \textbf{19.64}& 37.37& 38.15\\ \hline
    \end{tabular}
    }
    \label{tab:finetune2}
\end{table*}     

\subsubsection{Accuracy of Pseudo Labels}

With effective trajectory and endpoint embeddings, we exploit a small labeled set $\tilde{C}$ to assign pseudo labels to the larger crowdsourced dataset $C$, using a similarity threshold $\delta$. Note that the endpoints in $C$ are evenly distributed. \Tab{tab:pseudo-labeling} reports the results for Field 1. With 1K labeled samples and $\delta = 0.9$, the localization errors are $2.45$m (CDF68) and $4.93$m (CDF95), with $3.3\%$ of trajectories left unlabeled. Reducing the threshold to $\delta = 0.8$ increases errors to $2.91$m and $6.63$m, respectively, but nearly all trajectories are labeled (only $0.01\%$ dropped). This suggests that label quality matters more than quantity, and a higher threshold yields more accurate pseudo labels.

Using only 0.5K labeled samples, we observe the same trend: $\delta = 0.9$ yields better accuracy than $\delta = 0.8$ across both CDF68 and CDF95. However, overall performance declines compared to the 1K case, indicating that while quality is critical, quantity still plays a supporting role. Field 2 exhibits similar patterns, but with degraded accuracy due to its larger area and sparser AP density.

\subsubsection{Effectiveness of the Two-Stage Design}
\label{sec:EXP1-compare}

We evaluate the two-stage framework against an endpoint-only baseline (only cut-and-flip augmentation) that skips trajectory embedding and directly learns endpoint representations. As shown in \Tab{tab:pseudo-labeling}, the two-stage model achieves a strong localization accuracy with 0.5K labeled samples, which is $2.84$m (CDF68) and $5.39$m (CDF95) in $\delta = 0.9$. In contrast, the endpoint-only model performs poorly regardless of threshold, with errors exceeding $28$m (CDF68) and $36$m (CDF95) in both $\delta = 0.8$ and $\delta = 0.9$. These results highlight the importance of trajectory-level embedding. Without it, the model fails to capture discriminative movement patterns, leading to large localization errors.


\subsection{Comparisons}
\label{sec:EXP2}

We compare our method against several baselines: (i) KNN, (ii) LSTM trained on the labeled set $\tilde{C}$, (iii) NCP, our architecture without pretraining and without using $C$, and (iv) FULL, our complete model. Only FULL takes advantage of the unlabeled crowd-sourced dataset $C$ through pseudo labeling.

\Tab{tab:finetune} shows results in field 1 using $\tilde{C} = 0.5$K. Even without $C$, our FULL model, with pretraining on $\tilde{C}$, achieves better accuracy than LSTM and NCP (e.g., $6.7$m CDF68 vs. $7.16$m and $8.37$m). Adding 10K pseudo-labeled samples from $C$ improves accuracy to $5.77$m (CDF68), and up to $4.16$m with 200K samples. Varying $\delta$ controls the quality–quantity trade-off where $\delta=0.9$ produces the best result, while $\delta=0.99$ limits usable data and degrades accuracy.

The results in field 2 (\Tab{tab:finetune2}) show larger errors due to lower label density and sparser AP coverage. However, the trend remains consistent where pseudo-labeled data from $C$ substantially improve localization. Optimal performance is achieved again with $\delta=0.9$.

\section{Conclusions} \label{Sec:conclusion}

We have proposed a two-stage semi-self supervised training for learning representations of unlabeled crowdsourced data. Separating trajectory embedding and endpoint embedding has been proved helpful in predicting the endpoint of a WiFi trajectory. A novel cut-and-flip mechanism is proposed to self-learn the representations of unlabeled crowdsourced trajectories. 
The framework has been tested on small-scale datasets. 
The results could greatly reduce laborious data labeling costs.
Future work can be directed to more extensive tests and the next-generation WiFi technologies.

\thanks{
Acknowledgement: Y.-C. Tseng's research was sponsored by NSTC grants 
113-2639-E-A49-001-ASP and 
113-2634-F-A49-004. 
}


\small
\bibliographystyle{IEEEtran}
\bibliography{main}

\end{document}